\pdfoutput=1
\documentclass[11pt]{article}

\usepackage[preprint]{acl}

\usepackage{times}
\usepackage{latexsym}
\usepackage{algorithm}
\usepackage{algpseudocode}
\usepackage{amsmath}
\usepackage{amsfonts}  
\usepackage{amssymb}
\usepackage{multirow}
\usepackage{graphicx}
\usepackage{subcaption}
\usepackage{makecell}
\usepackage{booktabs}
\usepackage{array}
\usepackage{url}
\usepackage{dsfont}
\usepackage{xspace}
\usepackage{tcolorbox}
\usepackage{pifont}
\usepackage{xcolor}
\usepackage[T1]{fontenc}
\usepackage[utf8]{inputenc}

\usepackage{microtype}

\usepackage{inconsolata}

\usepackage{graphicx}

\tcbuselibrary{listings, breakable, skins, theorems}
\newcommand{\methodName}{\textsc{Icon$^{2}$}\xspace}

\renewcommand{\textcolor}[2]{#2} % 注释了这句话可以显示有颜色的版本

\newtcolorbox{prompt}[1]{
    enhanced,
    drop shadow=black!5!white,
    left=4mm,
    right=4mm,
    top=1mm,
    bottom=1mm,
    boxsep=0mm,
    rounded corners,
    title=#1,
    fontupper=\normalsize\linespread{1}\fontfamily{lmr}\selectfont,
}

\title{\methodName{}: Aligning Large Language Models Using Self-Synthetic\\Preference Data via Inherent Regulation}

\author{
    \textbf{Qiyuan Chen}$^{1,2}$\thanks{\, Equal Contribution.},
    \textbf{Hongsen Huang}$^{2}$\footnotemark[1],
    \textbf{Qian Shao}$^{1}$,
    \textbf{Jiahe Chen}$^{1}$,
    \textbf{Jintai Chen}$^{3}$\\
    \textbf{Hongxia Xu}$^{1}$,
    \textbf{Renjie Hua}$^{2,4}$,
    \textbf{Chuan Ren}$^{2}$\footnotemark[2],
    \textbf{Jian Wu}$^{1}$\thanks{\, Corresponding Author.}\\
    $^{1}$Zhejiang University, Zhejiang, China 
    $^{2}$Soochow Securities Co.,Ltd., Jiangsu, China \\
    $^{3}$HKUST(GZ), Guangdong, China
    $^{4}$Nanjing University, Jiangsu, China \\
    \texttt{qiyuanchen@zju.edu.cn},
}

\begin{document}
\maketitle
\begin{abstract}
Large Language Models (LLMs) require high quality preference datasets to align with human preferences. However, conventional methods for constructing such datasets face significant challenges: reliance on pre-collected instructions often leads to distribution mismatches with target models, while the need for sampling multiple stochastic responses introduces substantial computational overhead.
In this work, we explore a paradigm shift by leveraging inherent regulation of LLMs' representation space for efficient and tailored preference dataset construction, named \methodName{}.
Specifically, it first extracts layer-wise direction vectors to encode sophisticated human preferences and then uses these vectors to filter self-synthesized instructions based on their inherent consistency. During decoding, bidirectional inherent control is applied to steer token representations, enabling the precise generation of response pairs with clear alignment distinctions.
Experimental results demonstrate significant improvements in both alignment and efficiency. Llama3-8B and Qwen2-7B achieve an average win rate improvement of 13.89\% on AlpacaEval 2.0 and 13.45\% on Arena-Hard, while reducing computational costs by up to 48.1\%.
\end{abstract}

\section{Introduction}

Large Language Models (LLMs) have demonstrated remarkable capabilities across various NLP tasks, with a key factor behind their success being the alignment of LLMs from human preferences\citep{achiam2023gpt,brown2020language,touvron2023llama}.
While numerous preference learning algorithms have been developed to enhance this alignment~\citep{ouyang2022training,rafailov2024direct}, their effectiveness critically depends on large-scale, high-quality human-annotated preference datasets, which are notoriously challenging to obtain.
Such datasets, typically structured as triplets of instructions, chosen responses (human-preferred output), and rejected responses (dispreferred output), impose significant collection costs due to the intensive human labor required for annotation~\citep{cui2023ultrafeedback,dong2024self}.

\begin{figure}[t]
  \centering
  \includegraphics[width=\linewidth]{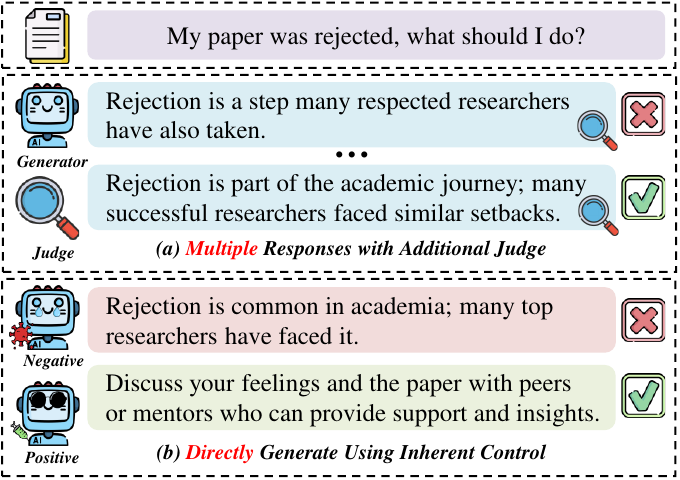}
  \caption{\textcolor{red}{Comparison of approaches: (a) Previous methods require multiple responses, while (b) \methodName{} can directly generate both chosen and rejected responses.}}
  \label{fig:compare}
  \vspace{-6mm}
\end{figure}

To address these challenges, researchers have explored using LLMs to construct preference datasets, such as substituting human preferences with advanced LLMs \citep{cui2023ultrafeedback,yuan2024self}, scoring responses with reward models ~\citep{xu2024magpie}, or refining outputs iteratively~\citep{dong2024self,cheng2024spar}.
Despite their effectiveness, these methods generate multiple responses for each instruction in a pre-collected dataset, which introduces two significant limitations.
First, the reliance on pre-collected instructions often results in preference datasets that lack customization for the specific characteristics of the target LLM. This limitation leads to distributional mismatches, which reduce alignment efficiency and generalization ability~\citep{xu2024magpie,yang2024self}, and may even result in catastrophic forgetting~\citep{huang2024mitigating}.
% Second, the uncontrollable stochastic variations in LLMs make it difficult to consistently control preference distinctions between chosen and rejected responses~\citep{dong2024self}. To mitigate this issue, a sufficient number of responses must be sampled for each instruction to ensure these distinctions, which in turn introduces unnecessary computational overhead.
\textcolor{red}{Second, the inherent stochasticity in LLM outputs makes it difficult to reliably control the qualitative distinction between chosen and rejected responses~\citep{dong2024self}. Consequently, multiple candidate responses must often be sampled and filtered for each instruction to ensure a meaningful preference gap, incurring substantial and often prohibitive computational overhead.}

Reflecting on these limitations, we find that existing methods predominantly rely on external randomness, overlooking the internal properties of LLMs themselves, which offer a more deterministic and structured way to encode sophisticated human preferences~\citep{zou2023representation,feng2024legend,liu2024ctrla}.
This realization leads us to ask: \emph{Could we instead regulate the inherent representation space of LLMs to integrate these preferences directly into generation?}
By shifting the focus to the inherent representation within LLMs, we propose \methodName{}, a unified framework that includes tailored instruction selection and precise token-level steering during decoding~\citep{liu2023aligning,zhang2024truthx,ji2024aligner}, thereby addressing the key challenges of customization, efficiency, and controllability in a systematic way.

Specifically, we first extract layer-wise direction vectors from the representation space of LLMs to capture sophisticated human preferences, such as honesty, harmlessness, and helpfulness. These vectors are derived using contrastive system prompts and aggregated through PCA to identify the most representative directions for each criterion.
Next, the model self-synthesizes a diverse set of instructions, which are then filtered based on their \underline{\textbf{i}}nherent \underline{\textbf{con}}sistency with the extracted preference directions, ensuring tailored customization with the target LLM's capabilities.
Finally, we employ bidirectional \underline{\textbf{i}}nherent \underline{\textbf{con}}trol to steer token representations during decoding, enabling the direct generation of response pairs with precise alignment differences, thereby eliminating the need for multiple responses.
Figure~\ref{fig:compare} illustrates the key distinctions between our approach and previous methodologies.

The experimental results demonstrate that \methodName{} not only enhances the alignment of LLMs with human preferences but also delivers significant computational efficiency. In particular, the Llama3-8B and Qwen2-7B models achieve notable improvements in length-controlled win rates, reaching 17.63 and 10.15 on AlpacaEval 2.0, and 13.7 and 13.2 on Arena-Hard, respectively. More importantly, \methodName{} achieves up to a 48.1\% reduction in computational costs compared to other baselines.

Our contributions can be summarized as follows: (1) We propose \methodName{}, a novel and systematic approach for efficiently constructing tailored preference datasets. (2) \methodName{} extracts direction vectors from the representation space of LLMs, utilizing inherent regulation for instruction filtering and precise response generation. (3) Experiments demonstrate superior alignment and efficiency of \methodName{}, achieving high performance with significantly fewer resources.
\section{Related Works}

\subsection{Preference Data Construction}

The construction of preference data typically relies on manual annotation~\citep{ouyang2022training,bai2022training,nakano2021webgpt} or advanced LLMs~\citep{cui2023ultrafeedback,ding2023enhancing}, such as GPT-4, to label different responses. 
To mitigate the substantial costs associated with these methods, there has been growing research interest in leveraging LLMs themselves to generate preference data.
Strategies include implementing reward models to select responses with higher rewards~\citep{zhang2024rest,tian2024toward}, using LLMs as judges to filter responses that better conform to human preferences~\citep{wang2024self,wu2024meta,yuan2024self} or utilizing self-play mechanisms to refine response quality~\citep{cheng2024spar,dong2024self,chen2024self}. 
Previous methods often generate multiple responses per instruction to ensure preference distinctions, but this introduces significant computational overhead and amplifies stochastic inconsistencies~\citep{dong2024self}. In contrast, our approach directly generates response pairs with precise alignments, eliminating the need for excessive sampling while maintaining efficiency and consistency.

\subsection{Synthetic Data for LLMs}

Synthetic data as an efficient method for constructing training data for LLMs has garnered considerable attention~\citep{tan2024large,long2024llms}. Previous approaches can generally be divided into two categories. One category is based on a seed-data paradigm, where methods typically rely on predefined seed instructions~\citep{honovich2023unnatural,wang2023self,xu2023wizardlm} or seed topics~\citep{li2024synthetic,gunasekar2023textbooks}, allowing strong LLMs to synthesize more diverse data based on these examples.
Another approach involves training specialized instruction synthesis models to generate diverse instructions~\cite{ding2024unleashing,dong2024self}. The fine-tuned models can generate a variety of instructions by sampling from a broad search space without the need for additional seed instructions or knowledge constraints.
Our approach directly constructs instruction data by prompting aligned LLMs with a pre-query template for sampling instructions~\cite{xu2024magpie}. Afterwards, we employed a novel inherent consistency filtering approach to select samples that are more tailored towards the target LLMs.
\section{\methodName{}: Aligning Large Language Models using Self-Synthetic Preference Data via Inherent Regulation}

In this section, we demonstrate how \methodName{} synthesizes datasets for preference optimization without requiring additional annotation or training. We begin by introducing the extraction of linear representation features in Section~\ref{sec:3.1}. These features are then utilized for both the selection of instructions, as detailed in Section~\ref{sec:3.2}, and the generation of responses, as described in Section~\ref{sec:3.3}.

\subsection{Linear Representation Feature Extraction}
\label{sec:3.1}

Building on the linear representation and superposition hypotheses~\citep{olah2023distributed,bricken2023towards,templeton2024scaling,zou2023representation}, our methodology extracts features that encode sophisticated human preferences from the representation space of LLMs. To achieve precise feature extraction~\citep{zou2023representation}, we deconstruct complex human preferences into fundamental criteria. Inspired by~\citet{liu2023aligning,tekin2024h}, we define the set of criteria as $\mathcal{C} = \{\text{honesty, harmlessness, helpfulness, general}\}$, \textcolor{red}{where the first three represent basic principles (referred to as 3H later), and the last one serves as an additional general standard to cover a wider range of human preferences (referred to as General later).}
To capture the directions of advanced human preferences, we manually design contrastive system prompts for each criterion. These prompts enable the extraction of features that distinguish between positive and negative alignment with the specified criteria. More information about the contrastive system prompts for each criterion can be found in the Appendix~\ref{appendix:contrastive-system-prompts}.

For each criterion $c \in \mathcal{C}$, we define positive and negative system prompts, denoted as $\mathcal{P}_{c}^+$ and $\mathcal{P}_{c}^-$, which align with and contradict the criterion, respectively. Given a dataset $\mathcal{D}_{\text{feat}} = \{ d_1, \dots, d_{|\mathcal{D}_{\text{feat}}|} \}$ containing $|\mathcal{D}_{\text{feat}}|$ instructions, we concatenate each instruction $d_i$ with the positive and negative system prompts to form complete inputs $\mathcal{P}_c^+ \oplus d_i$ and $\mathcal{P}_c^- \oplus d_i$. These inputs are then fed into the LLMs to obtain their corresponding feature representations.

\begin{figure}[t]
    \centering
    \includegraphics[width=0.95\linewidth]{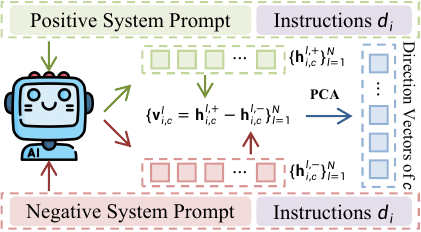}
    \caption{Direction Vector Computation: (a) Positive and negative system prompts $\mathcal{P}_{c}^+$ and $\mathcal{P}_{c}^-$ generate corresponding representations $\mathbf{h}_{i,c}^{l,+}$ and $\mathbf{h}_{i,c}^{l,-}$; (b) Contrastive vectors $\mathbf{v}_{i,c}^l$ are derived as the difference between positive and negative representations at each layer; (c) PCA extracts layer-wise direction vectors $\mathbf{u}_c^l$ for each criterion $c$.}
    \label{fig:vector}
    \vspace{-6mm}
\end{figure}

Considering the heterogeneous representation spaces across different layers of LLMs~\citep{chuang2023dola,sun2024transformer}, we extract the representation of the last token from each layer. 
\textcolor{red}{This choice is due to decoder architectures where causal attention~\citep{wang2023improving} ensures only the last token's representation at each layer has integrated the entire preceding sequence, thus serving as the layer's representation of the whole input.}
Formally, for each instruction $d_i$ and criterion $c \in \mathcal{C}$, we obtain representations $\{\mathbf{h}_{i,c}^{l,+}\}_{l=1}^{N}$ and $\{\mathbf{h}_{i,c}^{l,-}\}_{l=1}^{N}$ for the positive and negative system prompts, respectively. Here, $N$ denotes the total number of layers in the LLMs, and $\mathbf{h}_{i,c}^{l}$ represents the last token's representation at the $l$-th layer.

After obtaining the representations through contrastive system prompts, we propose to identify the direction vectors that characterize the target criterion $c$. As shown in Figure~\ref{fig:vector}, the contrastive vector at the $l$-th layer for instruction $d_i$ is formally defined as the vector difference between the representations of positive and negative inputs. Specifically, given the positive input $\mathcal{P}_c^+ \oplus d_i$ and negative input $\mathcal{P}_c^- \oplus d_i$, we compute their hidden state representations $\mathbf{h}_{i,c}^{l,+}$ and $\mathbf{h}_{i,c}^{l,-}$ respectively, then derive the contrastive vector as:

\begin{equation}
\mathbf{v}_{i,c}^l = \mathbf{h}_{i,c}^{l,+} - \mathbf{h}_{i,c}^{l,-}.
\label{eq:contrastive-vector}
\end{equation}

Following the methodology of \citet{zou2023representation}, we compute layer-wise direction vectors for criterion $c$ through contrastive vector aggregation and dimensionality reduction. Formally, for each layer $l \in [1, N]$, we first aggregate the contrastive vectors $\{\mathbf{v}_{i,c}^l\}_{i=1}^{|\mathcal{D}_{\text{feat}}|}$ across all instructions in dataset $\mathcal{D}_{\text{feat}}$. We then perform PCA on the aggregated vectors, where the first principal component $\mathbf{u}_c^l$ captures the maximal variance direction in the contrastive space. The final direction vectors of criterion $c$ is therefore defined as the layer-wise component set $\mathbf{u}_c = \{\mathbf{u}_c^l\}_{l=1}^N$.
% Sensitivity analysis shows the vector can robustly encodes human preferences (see Appendix~\ref{app:sensitivity_analysis_dfeat_narrative}).
\textcolor{red}{Appendix~\ref{app:sensitivity_analysis_dfeat_narrative} details a sensitivity analysis, indicating the representation's robustness to human preferences.}

\begin{figure*}[thb]
    \centering
    \includegraphics[width=0.95\linewidth]{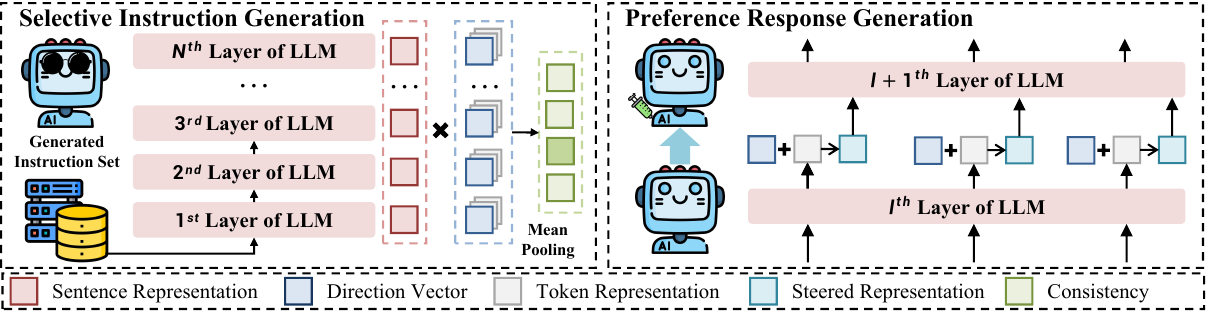}
    \caption{Framework for Instruction Filtering and Preference Response Generation. The process begins with a diverse set of synthesized instructions, which are filtered by measuring their consistency with predefined criteria using direction vectors extracted from contrastive representations. These vectors then guide the generation of preference response pairs through inherent control, where token representations are steered during decoding to produce chosen and rejected responses. This approach enables efficient and tailored dataset construction for preference optimization without additional annotations or multiple response generations.}
    \label{fig:framework}
    \vspace{-6mm}
\end{figure*}

\subsection{Selective Instruction Generation via Inherent Consistency}
\label{sec:3.2}

To produce a variety of tailored instructions, we employ a \emph{sample then select} paradigm~\citep{tan2024large}, which involves initially generating an extensive range of diverse instructions.
Previous methods typically rely heavily on prompt engineering and careful selection of initial instructions~\citep{xu2023wizardlm,wang2023self}, leading to a trend of decreasing synthetic data diversity as the size of the dataset increases, which is not conducive to scaling up the dataset. Thus, we aim to generate instructions without relying on seed instructions but rather by inputting the pre-query templates up to the position reserved for user messages, inspired by~\citet{xu2024magpie,ding2024unleashing}.

Specifically, for open-weight aligned language models, we design pre-query templates that match their predefined instruction formats. These auto-regressive LLMs, having been fine-tuned on data in similar formats, automatically generate appropriate instructions when provided with the template input. The generation process terminates upon producing an end-of-sequence token, ensuring instruction completeness. By repeating the above process multiple times, we obtain a diverse instruction set $\mathcal{D}_{\text{raw}}$ without requiring seed instructions or training.

After obtaining a diverse set of instructions, the next step is to filter out instructions that are more tailored to the target model, enabling it to achieve better results given the data size. To this end, we propose a method of instruction filtering using inherent consistency. Specifically, this step involves two aspects: on one hand, it is necessary to construct a high-quality and tailored subset of instructions; on the other hand, it is essential to identify the specific contribution of each instruction to the model's capabilities. For instance, the instruction "What is the model number of Xiaomi's latest phone?" primarily enhances the model's \emph{honesty}, whereas "Help me write a quick sort code" focuses more on improving the model's \emph{helpfulness}.

To effectively tailor instructions, we first evaluate their contributions across the predefined criteria $\mathcal{C}$. This evaluation helps identify which model capabilities each instruction is most likely to enhance. \textcolor{red}{For this purpose, we assess the alignment of an instruction with a specific criterion by measuring its \emph{inherent consistency}~\citep{zou2023representation}.} This involves comparing the criterion's feature direction in the representation space with the instruction's representations. We adopt this approach of monitoring inherent consistency, rather than relying on prompting or fine-tuning LLMs, as it more accurately reflect the model's internal understanding and alignment with the desired human preferences.

\textcolor{red}{Specifically, for a given instruction $d_i$ from the raw set $\mathcal{D}_{\text{raw}}$ and a criterion $c \in \mathcal{C}$, we utilize the extracted direction vectors $\mathbf{u}_c = \{\mathbf{u}_c^l\}_{l=1}^N$ (as detailed in Section~\ref{sec:3.1}) alongside the instruction's layer-wise representations $\mathbf{h}_i = \{\mathbf{h}_i^l\}_{l=1}^N$. The consistency score, $\text{consistency}_{i,c}$, which quantifies the alignment between instruction $d_i$ and criterion $c$, is then computed as the mean-pooled dot product across all $N$ layers:}
\begin{equation}
\text{consistency}_{i,c} = \mathtt{meanpool}\left( \left[ {\mathbf{h}_{i}^{l}}^{\top} \cdot \mathbf{u}_{c}^l \right]_{l=1}^{N} \right).
\label{eq:consistency-score}
\end{equation}
After computing $\text{consistency}_{i,c}$ values for all instruction-criterion pairs, we aim to assign a single, representative score to each instruction that reflects its overall potential for alignment. Since a higher consistency score indicates a stronger alignment between an instruction and a particular criterion, we define the final inherent consistency score for an instruction $d_i$, denoted as $\text{consistency}_i$, to be its maximum consistency value achieved with any criterion in $\mathcal{C}$. This approach ensures that the score captures the instruction's most prominent alignment with the defined capabilities. Formally, this is expressed as:
\begin{equation}
    \text{consistency}_{i}  = \max_{c \in \mathcal{C}} \, \text{consistency}_{i,c} .
\end{equation}
\textcolor{red}{This procedure results in a set of final inherent consistency scores, $\mathcal{A} = \{ \text{consistency}_i \}_{i=1}^{|\mathcal{D}_{\text{raw}}|}$, for all instructions present in $\mathcal{D}_{\text{raw}}$. These scores then serve as the basis for ranking or applying a threshold to filter the instructions. The ultimate goal is to curate a more tailored and high-quality subset, $\mathcal{D}_{\text{filt}}$, which comprises instructions demonstrating a strong alignment with only one specified criteria.}

\subsection{Preference Response Generation via Inherent Control}
\label{sec:3.3}

Datasets employed for preference optimization are generally structured as triplets, comprising an instruction, a chosen response, and a rejected response. Consequently, once a diverse and tailored collection of instructions is acquired, corresponding chosen and rejected responses should be generated for each instruction.

Previous methods often generate multiple responses for a single instruction, typically produced by different models and selected using reward models or advanced LLMs for labeling preferences~\citep{ouyang2022training,cui2023ultrafeedback,yuan2024self}. However, this approach introduces significant challenges. The variability in model capabilities can obscure subtle preference distinctions, while the need for distinct differences between chosen and rejected responses requires excessive sampling, leading to high computational costs and inefficient data utilization~\citep{dong2024self}. Additionally, reliance on reward models or advanced LLMs for annotation further increases complexity without ensuring consistent alignment with human preferences.
\textcolor{red}{While alternative strategies for generating preference pairs, such as using distinct positive and negative system prompts, might appear computationally efficient, they are prone to critical issues like reward hacking and lack of fine-grained control, rendering them impractical (See Appendix~\ref{app:direct_prompting_comparison} for more details).}

Therefore, we propose using inherent control to generate preference response pairs, which not only exhibit clear distinctions but also do not require multiple responses, thus enhancing the efficiency of constructing preference optimization datasets. \textcolor{red}{Crucially, for each instruction $d_i \in \mathcal{D}_{\text{filt}}$, this inherent control is guided by the specific criterion $c^* \in \mathcal{C}$ that yielded the highest consistency score $\text{consistency}_{i}$ for that instruction, as determined in Section~\ref{sec:3.2}. According to the superposition hypothesis~\citep{templeton2024scaling,liu2023aligning,ilharco2022editing}, aligning LLMs with this specific criterion $c^*$ can be enhanced by modifying token representations during decoding to steer the model's outputs closer to the direction that embodies this criterion. Specifically, for this identified criterion $c^*$, we can derive its direction $\mathbf{u}_{c^*} = \{\mathbf{u}_{c^*}^l\}_{l \in \mathcal{L}_{c^*}}$ (where $\mathcal{L}_{c^*} \subseteq [1,\dots,N]$ denotes the subset of controlled layers) from Equation~\ref{eq:contrastive-vector}. Let $\mathbf{Z}_k = \{\mathbf{z}_{k}^l\}_{l\in \mathcal{L}_{c^*}}$ represent the token representations for the $k$-th token at these controlled layers. We then apply a linear combination function for preference steering:}
\begin{equation}
\begin{aligned}
    \hat{\mathbf{Z}}_{k,c^*}&=\mathbf{Z}_{k}+\gamma_{c^*}\cdot\mathbf{u}_{c^*} \\
    &=\{\mathbf{z}_{k}^l+\gamma_{c^*}\cdot\mathbf{u}_{c^*}^l \; | \; \forall \; l \in \mathcal{L}_{c^*} \},
\label{eq:preference-steering}
\end{aligned}
\end{equation}
\textcolor{red}{In Equation~\ref{eq:preference-steering}, the coefficient $\gamma_{c^*}$ controls the steering intensity along $\mathbf{u}_{c^*}$ within the selected layers $\mathcal{L}_{c^*}$. To synthesize the preference pair $(r^{\text{chosen}}, r^{\text{rejected}})$ for each instruction, $r^{\text{chosen}}$ is generated using positive steering and $r^{\text{rejected}}$ using negative steering. This method requires \emph{exactly two generation passes} per instruction. As illustrated in Figure~\ref{fig:framework}, preference steering is applied layer-specifically and token-by-token during generation, offering a simple yet effective approach with less impact on inference costs.}
\section{Experiments}

In this section, we present our experimental results to answer the following question:
\begin{itemize}\vspace{-2mm}
    \item[$\circ$] Does \methodName{} improve the alignment of LLMs across various LLMs? (Section~\ref{exp:alpaca}, Table~\ref{tab:main_res})\vspace{-3mm} 
    \item[$\circ$] Is \methodName{} also effective to improve overall LLM's capability? (Section~\ref{exp:mtbench}, Table~\ref{tab:mtbench})\vspace{-3mm}
    \item[$\circ$] Can \methodName{} generate diverse and tailored instructions? (Section~\ref{exp:ins}, Table~\ref{tab:ins})\vspace{-3mm}
    \item[$\circ$] Does \methodName{} save the cost of preference data construction? (Section~\ref{exp:cost}, Table~\ref{tab:cost})\vspace{-3mm}
    \item[$\circ$] How do the hyperparameters introduced by \methodName{} affect model performance? (Section~\ref{exp:hyper}, Figure~\ref{fig:hyper})\vspace{-3mm}
\end{itemize}

% In this section, we present our experimental results to answer the following question:
% \begin{itemize}[leftmargin=5.5mm,topsep=0pt]\vspace{-0.04in}
%     \item[$\circ$] Does \name{} improve the alignment of LLMs only using a small amount of human-labeled preference data? (Table~\ref{tab:main_result}, Figure \ref{fig:res_no_seed}) \vspace{-0.04in}
%     \item[$\circ$] Does the proposed method outperform other preference labeling methods?  (Table~\ref{tab:judgments}, Figure \ref{fig:res_iteration}) \vspace{-0.04in}
%     \item[$\circ$] Is \name{} generalizable across various LLMs? (Table~\ref{tab:phi}) \vspace{-0.04in}
%     \item[$\circ$] Is \name{} also effective to improve overall LLM's capability? (Table~\ref{tab:mt_bench_result}) %\vspace{-0.04in}
% \end{itemize}

\subsection{Experimental Setups}
\paragraph{Models.}

We performed preference optimization on Qwen2-7B and Llama3-8B Base models, starting from supervised fine-tuned versions like \citet{meng2024simpo,dong2024self}. Both models were fine-tuned on the UltraChat-200k dataset using the LLaMA-Factory pipeline \citep{zheng2024llamafactory}.\footnote{\url{https://github.com/hiyouga/LLaMA-Factory}}.

\paragraph{Baselines.}

% In this study, we employ initial supervised fine-tuned models as baselines, along with those optimized by data derived from diverse preference construction methods, encompassing both manual collection and preference judgment techniques.
% We acknowledge UltraFeedback preference data~\citep{cui2023ultrafeedback} as manually collected from six high-quality datasets and various models, with preferences annotated by GPT-4~\citep{achiam2023gpt}.
% For preference judgment approaches, we utilize two methods: Sampling-Ranking and Self-Rewarding. The Sampling-Ranking method, similar to~\citet{meng2024simpo} and~\citet{dong2024self}, involves LLMs sampling five responses for each instruction. Subsequently, the same scoring models, namely ArmoRM-Llama3-8B-v0.1\footnote{\url{https://huggingface.co/RLHFlow/ArmoRM-Llama3-8B-v0.1}}~\citep{wang2024interpretable}, selects the highest and lowest-scoring responses as the chosen and rejected responses, respectively. 
% In the Self-Rewarding method~\citep{yuan2024self}, we generate preference data based on the model's self-assessed rewards using the LLM-as-a-Judge prompting~\citep{bai2022training}. 
% As for the Self-Refine method~\citep{kim2024aligning,dong2024self}, we first use LLMs to sample three responses for each instruction, then use the Self-Refine prompt to get the chosen response and randomly select the pre-refine response as the reject response.
% Details are presented in Appendix~\ref{app:exp_detail}.

Our study uses initial SFT models as baselines, alongside models optimized with preference data from various methods. We utilize UltraFeedback~\citep{cui2023ultrafeedback}, a manually collected dataset with preferences annotated by GPT-4~\citep{achiam2023gpt}. For preference judgment, we employ Sampling-Ranking, similar to~\citet{meng2024simpo}, where LLMs sample five responses per instruction, and ArmoRM-Llama3-8B-v0.1~\citep{wang2024interpretable} selects the chosen and rejected responses with reward scores. We also use the Self-Rewarding method~\citep{yuan2024self}, generating preference data based on the model's self-assessed rewards via LLM-as-a-Judge prompting~\citep{bai2022training}. Additionally, the Self-Refine method~\citep{kim2024aligning,dong2024self} involves LLMs sampling three responses, using a Self-Refine prompt for the chosen response, and randomly selecting a pre-refine response as rejected. More details are in Appendix~\ref{app:exp_detail}.

\paragraph{Evaluations.}

We evaluate the model alignment performance on AlpacaEval 2.0~\citep{alpaca_eval} and Arena-Hard~\citep{arenahard2024}, and overall capabilities on MT-Bench~\citep{zheng2023judging}. More details about evaluation datasets can be found in the Appendix~\ref{app:eval-datasets}.
To further enhance the robustness of the verification, we conducted a leakage analysis on the synthetic preference dataset. Details are presented in Appendix~\ref{app:data-leakage}.

\paragraph{Implementation Details.}
 
For all experiments, we performed one epoch of offline DPO with a fixed $\beta = 0.1$. The global batch size was set to 128, and the learning rate was $5 \times 10^{-7}$. For the hyperparameters introduced by our method, we set $\gamma_c=0.1$ for the chosen response and $\gamma_c=-0.05$ for the rejected response. For all models, the control layer interval is set to $[10,20]$.
$\mathcal{D}_{\text{raw}}$ contains 1M diverse English instructions, $\mathcal{D}_{\text{filt}}$ contains 100K filtered instructions, of which 98K are for training and 2K for validation.
$\mathcal{D}_{\text{feat}}$ is composed of 1024 samples randomly selected from the Alpaca dataset~\cite{alpaca}.
Additional implementation details are available in Appendix~\ref{app:impl_detail}.

\setlength{\tabcolsep}{5pt}
\begin{table*}[!t]
\centering
\resizebox{0.95\textwidth}{!}{
\begin{tabular}{lcccccc}
\toprule
\multirow{3}{*}{\textbf{Data Construction}} & \multicolumn{3}{c}{\textbf{Llama3-Base (8B)}} & \multicolumn{3}{c}{\textbf{Qwen2-Base (7B)}} \\ 
\cmidrule(lr){2-4}\cmidrule(lr){5-7}
& \multicolumn{2}{c}{\textbf{AlpacaEval 2.0}} & \multicolumn{1}{c}{\textbf{Arena-Hard}} & \multicolumn{2}{c}{\textbf{AlpacaEval 2.0}} & \multicolumn{1}{c}{\textbf{Arena-Hard}}\\ 
\cmidrule(lr){2-3}\cmidrule(lr){4-4}\cmidrule(lr){5-6}\cmidrule(lr){7-7}
& \bf LC (\%) & \bf WR (\%) &  \bf WR (\%) & \bf LC (\%) & \bf WR (\%) &  \bf WR (\%)  \\
\midrule
SFT & 5.59 & 3.11 & 2.7 & 9.95 & 4.53 & 3.8 \\
\midrule
Manual Collection & 12.68 & 7.90 & 9.5 & 16.26 & 10.00 & 11.2 \\
\midrule
Sampling-Ranking & 16.52 & 10.43 & 13.9 & 17.24 & 11.42 & 15.3\\
Self-Rewarding & 16.02 & 10.19 & 13.5 & 16.46 & 10.37 & 14.7\\
Self-Refine & 18.38 & 12.80 & 14.2 & 17.39 & 11.80 & 16.2 \\
\midrule
\methodName{} (General) & 16.07 & 10.12 & 13.4 & 17.24 & 11.74 & 14.9 \\
\methodName{} (3H) & 18.63 & 15.22 & 14.5 & 19.13 & 12.17 & 15.6 \\
\methodName{} (General+3H) & \bfseries23.22 & \bfseries16.40 & \bfseries16.4 & \bfseries20.00 & \bfseries12.73 & \bfseries17.0 \\
\bottomrule
\end{tabular}
}
\caption{Performance comparison of different preference dataset construction methods on AlpacaEval 2.0 and Arena-Hard benchmarks. The metrics reported include length-controlled win rates (LC) and raw win rates (WR) for two base models: Llama3-Base (8B) and Qwen2-Base (7B). These results highlight the effectiveness of the proposed approaches in enhancing model performance across diverse evaluation settings.}
\label{tab:main_res}
\vspace{-5mm}
\end{table*}

\subsection{Evaluation on AlpacaEval 2.0 and Arena-Hard}\label{exp:alpaca}

We compare the instruction-following and human preference alignment capabilities on AlpacaEval 2.0~\citep{alpaca_eval} and Arena-Hard~\citep{arenahard2024} in Table~\ref{tab:main_res}.
Compared to the initial model after SFT, \methodName{} can significantly improve the win rate on different benchmarks.
On AlpacaEval 2.0, Llama3-Base achieved the highest increase of 17.63\% in length control win rate and 13.29\% in raw win rate. Similarly, Qwen2 achieved the highest increases of 10.05\% and 8.2\%, respectively.
In the more challenging Arena-Hard setting, \methodName{} also achieved the highest improvements of 13.7 and 13.2, respectively.
Moreover, the setting of General+3H always achieves the best performance, surpassing all conventional baseline methods, indicating that fine-grained attribution through inherent consistency for each instruction, followed by targeted inherent control, can effectively improve the quality of responses.
More results using different model sizes can be found in the Appendix.
% \setlength{\tabcolsep}{5pt}
% \begin{table}[t]
%         \centering  
%         \setlength{\tabcolsep}{2pt}
%         \resizebox{0.99\textwidth}{!}{
%         \begin{tabular}{lcccc}  
%             \toprule  
%             \multirow{2}{*}{\bf Data Construction} & \multicolumn{2}{c}{\textbf{Mistral-Base}} & \multicolumn{2}{c}{\textbf{Llama3-Base}} \\  
%             \cmidrule(lr){2-3} \cmidrule(lr){4-5}  
%             & \bf Turn 1 & \bf Turn 2 & \bf Turn 1 & \bf Turn 2 \\  
%             \midrule  
%             SFT & 6.04 & 5.65 & 6.55 & 5.36 \\ 
%             \midrule
%             Manual Collection  & 6.73   & \bf 6.82   & 7.29 & 7.00       \\
%             Sampling-Ranking  & 6.83 & 6.18 & 7.06 & 6.99 \\
%             Self-Rewarding  & 6.71 & 6.63 & 7.30 & 7.28\\
%             Self-Rewarding  & 6.53 & 6.41 & 6.99 & 6.65 \\  
%             Self-Rewarding \textit{Iter2} & 6.66 & 6.65 & 7.34 & 7.30 \\  
%             Self-Rewarding \textit{Iter3} & \bf 6.86 & \bf 6.82 & 7.34 & \bf 7.34 \\  
%             Self-Rewarding \textit{Iter4} & 6.73 & 6.69 & \bf 7.43 & 7.04 \\  
%             \bottomrule
%         \end{tabular} 
%         }
%         \caption{Multi-turn evaluation on MT-Bench. An asterisk (*) denotes the best score across multiple iterations. For Sampling-Ranking, Llama's best is from iteration 4 and Mistral's from iteration 3. For Self-Rewarding, both are from iteration 3.  \methodName{} progressively enhances the multi-turn instruction-following capabilities of LLMs.
%         \label{tab:mtbench}
% \end{table}
\setlength{\tabcolsep}{5pt}
\begin{table}[h]
\centering
\resizebox{0.99\linewidth}{!}{
\begin{tabular}{lcccc}
    \toprule  
    \multirow{2}{*}{\bf Data Construction} & \multicolumn{2}{c}{\textbf{Llama3-Base}} & \multicolumn{2}{c}{\textbf{Qwen2-Base}} \\  
    \cmidrule(lr){2-3} \cmidrule(lr){4-5}  
    & \bf Turn 1 & \bf Turn 2 & \bf Turn 1 & \bf Turn 2 \\  
    \midrule  
    SFT & 6.54 & 5.76 & 6.80 & 6.00 \\
    \midrule
    Manual Collection  & 7.09 & 6.44 & 7.15 & 6.11 \\
    Sampling-Ranking  & 6.95 & 6.00 & 7.32 & 6.39 \\
    Self-Rewarding  & 6.98 & 6.14 & 7.30 & 6.31 \\  
    Self-Refine  & 7.10 & 6.56 & 7.46 & 6.56 \\  
    \methodName{}(General)  & 6.95 & 6.20 & 7.51 & 6.79 \\  
    \methodName{}(3H)  & 7.13 & 6.83 & 7.85 & 6.83 \\  
    \methodName{}(General+3H)  & 7.38 & 6.76 & 7.68 & 7.11 \\  
    \bottomrule
\end{tabular}
}
\caption{Multi-turn evaluation results on MT-Bench comparing different preference data construction methods for Llama3-Base and Qwen2-Base models.}

\label{tab:mtbench}
\vspace{-6mm}
\end{table}

\subsection{Evaluation on MT-Bench}\label{exp:mtbench}

To further validate the improvements of \methodName{} in overall capabilities and multi-turn dialogue, we conducted evaluations on MT-Bench, with the results shown in Table~\ref{tab:mtbench}.
\methodName{}(General+3H) achieves the strongest overall performance on both base models under multi-turn evaluation which indicate that \methodName{} enhances not only first-turn performance, with an increase of over 0.86 points, but also the second turns, with an increase of over 1.05 points.
Similarly, General+3H also achieved the best performance, demonstrating its generalization capability.
The specific scores for different instruction types can be found in the Appendix~\ref{app:mtbench}.

\setlength{\tabcolsep}{2pt}  
\begin{table}[t]
    \centering
    \resizebox{0.99\linewidth}{!}{
        \begin{tabular}{lccccc}  
            \toprule
            \multirow{2}{*}{\textbf{Data}} & \multirow{2}{*}{\textbf{Filter}} & \multicolumn{2}{c}{\textbf{\methodName{}}} & \multicolumn{2}{c}{\textbf{Sampling-Ranking}} \\  
            \cmidrule(lr){3-4} \cmidrule(lr){5-6}  
            & & \bf LC (\%) & \bf WR (\%) & \bf LC (\%) & \bf WR (\%) \\  
            \midrule  
            \multirow{2}{*}{SI} & \ding{55} & 11.1 & 7.0 & 10.2 & 6.7 \\ 
                                & \ding{51} & 11.6 & 8.9 & 10.8 & 8.6 \\ \midrule  
            \multirow{2}{*}{MC} & \ding{55} & 12.8 & 9.7 & 10.4 & 7.5 \\             
                                & \ding{51} & 13.9 & 10.6 & 12.0 & 9.3 \\ \midrule  
            \multirow{2}{*}{T2} & \ding{55} & 15.7 & 10.7 & 12.8 & 8.7 \\             
                                & \ding{51} & 16.8 & 12.3 & 14.2 & 9.7 \\ \midrule  
            \multirow{2}{*}{\methodName{}} & \ding{55} & 17.1 & 12.0 & 15.0 & 10.1 \\  
                                           & \ding{51} & \bf18.0 & \bf13.3 & 16.0 & 11.2 \\  
            \bottomrule
        \end{tabular}
    }
    \caption{Performance comparison on AlpacaEval 2.0. Instructions derived from Manual Collection (MC), Self-Instruct (SI), Tulu V2 (T2), and the proposed \methodName{}. In the filter column, \ding{51} indicates filtering with inherent consistency, while \ding{55} indicates random selection.}
    \vspace{-6mm}
    \label{tab:ins}
\end{table}

\subsection{The Impact of Original Instructions and Filtering Method}\label{exp:ins}

To demonstrate that \methodName{} can generate diverse and customized instructions, we compare the instructions self-synthesized by \methodName{} with manual collected instructions~\citep{cui2023ultrafeedback}, Self-Instruct~\citep{wang2023self}, and Tulu V2~\citep{ivison2023camels}. In addition, we also analyze the performance differences caused by various filtering methods.
For each instruction construction method, we sampled 20k instructions using random selection or filtering with inherent consistency, constructed response pairs via \methodName{} and Sampling-Ranking, and performed DPO on Llama3-8B.
As presented in Table~\ref{tab:ins}, both the \methodName{} and Sampling-Ranking methods demonstrate that the self-synthesized instructions result in better-aligned models, validating the quality of these instructions.
This improvement may be attributed to two key factors: 1) eliminating seed data dependency by leveraging pre-query templates to generate diverse instructions, and 2) generating high-quality instructions that scale with LLM advancements. Moreover, filtering instructions based on inherent consistency improves model performance, demonstrating the effectiveness of our method.

\subsection{Cost Analysis}\label{exp:cost}

Our method introduces slight computational overhead in preference data construction. While each Transformer layer in LLMs has $\mathcal{O}(n^2d + nd^2)$ complexity, \methodName{} introduces only an additional $\mathcal{O}(nd)$ computational overhead through simple vector operations on the representations from specific layers.
As shown in Table~\ref{tab:cost}, \methodName significantly reduces GPU hours by eliminating the need for generating multiple responses. \textcolor{red}{And the preference annotation stage, consisting of efficient direction extraction (requiring only 2.2 GPU hours) and consistency calculation (8.6 GPU hours for 100k representations), contributes to a significantly more efficient preference data construction process.}
The approach not only achieves better alignment but also maintains lower computational costs compared to conventional methods.
\setlength{\tabcolsep}{5pt}
\begin{table}[t]
\centering
\resizebox{0.95\linewidth}{!}{
\begin{tabular}{lccc}
\toprule
\multirow{2}{*}{\textbf{Method}} 
& \multicolumn{2}{c}{\textbf{GPU Hours (h)}} 
& \multirow{2}{*}{\textbf{Cost (\$)}} \\
\cmidrule(lr){2-3}
& \textbf{Response} & \textbf{Preference} &  \\
\midrule
Sampling-Ranking & 123.8 & 7.2 & 269.9 \\
Self-Rewarding & 123.8 & 15.6 & 287.2 \\
Self-Refine & 75.4 & 26.3 & 209.5 \\
\midrule
\methodName{} & \textbf{61.6} & \textbf{10.8} & \textbf{149.1} \\
\bottomrule
\end{tabular}
}
\caption{Comparison of computational costs across methods on Llama3-8B. \methodName{} reduces GPU hours and cost by optimizing response generation and preference annotation, achieving better alignment with lower overhead compared to conventional approaches.}
\label{tab:cost}
\vspace{-6mm}
\end{table}

% Compared to other methods, our approach achieves better alignment while maintaining lower computational costs. Although~\methodName{} introduces a slight additional computational overhead of $\mathcal{O}(nd)$ in preference data construction through simple vector operations on representations from specific layers (significantly less than the $\mathcal{O}(n^2d + nd^2)$ complexity of Transformer layers), it markedly reduces GPU hours by eliminating the need to generate multiple responses. Its efficient preference annotation stage further cuts costs, comprising direction extraction (e.g., acquiring four directions requires a total of only 2.2 GPU hours, and extracting 100k representations takes about 8.6 GPU hours) and consistency calculation, which can be performed on the CPU.
% \input{tables/cost}

\subsection{The Impact of Hyperparameters on Model Performance}\label{exp:hyper}

\begin{figure}[t]
    \centering
    \includegraphics[width=\linewidth]{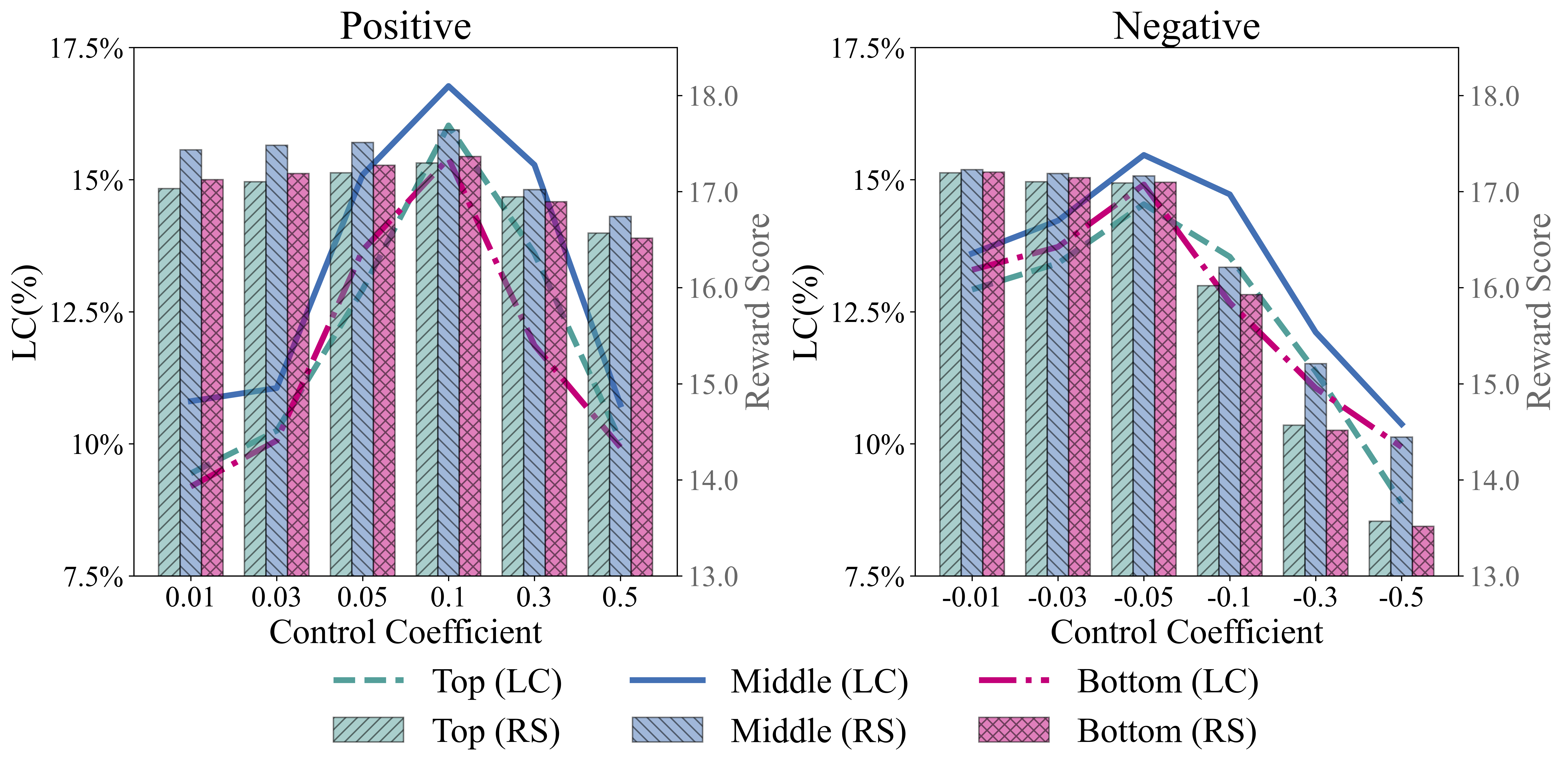}
    \caption{\textcolor{red}{Performance comparison of different controlled layers and $\gamma_c$ on Alpacaeval 2.0. Different colors of lines represent different layer intervals, where the middle layer always achieves the best alignment effect.}}
    \label{fig:hyper}
    \vspace{-6mm}
\end{figure}

To demonstrate hyperparameter impact, we created a preference dataset from 20k instructions, varying layer ranges (top: $[2, 12]$, middle: $[10, 20]$, bottom: $[20, 30]$) and control coefficient $\gamma_c$, and optimized Llama3-8B. We also present the mean reward score (RS) of responses across all settings. Figure~\ref{fig:hyper} shows the LC on Alpacaeval 2.0 and RS.
\textcolor{red}{Based on these results, we derive an efficient hyperparameter tuning method that requires only a small number of responses for each setting to obtain the optimal hyperparameters, without the need for DPO. More details can be found in Appendix~\ref{sec:appendix_hyperparameter_selection}.
}

% To illustrate the impact of hyperparameters on model performance, we selected 20k instructions and constructed a preference dataset by varying the layer ranges and control coefficient $\gamma_c$. Specifically, we divided the layers into three intervals: top layers in the range $[2, 12]$, middle layers in the range $[10, 20]$, and bottom layers in the range $[20, 30]$. Using these configurations, we performed preference optimization on Llama3-8B. 
% \textcolor{red}{In addition, we also present the mean of the reward scores (RS) of responses across all settings.}
% The LC on Alpacaeval 2.0 and RS are shown in Figure~\ref{fig:hyper}.
% Based on the experiments and conclusions in this section, we can derive an efficient hyperparameter tuning method without DPO. Details can be found in Appendix~\ref{sec:appendix_hyperparameter_selection}.

\subsubsection{Impact of Controlled Layers}

Early layers of neural networks struggle to capture full input text representation, while deeper layers are more task-specific. Studies suggest middle layers excel at capturing concept-related information~\citep{zou2023representation,liu2023aligning}. Modifying these layers introduces meaningful preference variations without significantly impacting output quality, making them ideal for preference dataset construction. This phenomenon is observed across all $\gamma_c$ in both positive and negative control, highlighting their utility in preference optimization. \textcolor{red}{Furthermore, responses generated using the middle layer achieved the highest reward values, reinforcing these findings. Therefore, only a small number of responses are needed to determine the selected number of layers via RS.}

% The early layers of neural networks are less capable of capturing the representation of the entire input text, whereas the layers closer to the top are more task-specific. Previous studies have demonstrated that representations extracted from the middle layers are more effective in capturing concept-related information~\citep{zou2023representation,liu2023aligning}. Notably, modifications to the middle layers introduce meaningful preference variations without significantly affecting output quality, making them ideal for constructing preference datasets. This phenomenon can be observed across all $\gamma_c$ in both positive and negative control, further highlighting their utility in preference optimization. The responses obtained using the intermediate layer also showed the highest reward values, similarly revealing similar conclusions.

\subsubsection{Impact of $\gamma_c$}

% The coefficient $\gamma_c$ controls the strength of preference steering. For both positive and negative control, a larger $\gamma_c$ does not necessarily lead to better results, as a high $\gamma_c$ introduces additional noise, whereas a low $\gamma_c$ can result in inadequate control.
% However, the optimal $\gamma_c$ for negative control is smaller, suggesting that the model is more sensitive to negative control.
% At the same time, when changing hyperparameters, positive control has a greater impact on the final result, highlighting the importance of positive responses' quality in preference optimization.
% And in different layer intervals, the impact of $\gamma_c$ on performance shows similar variation trends.
$\gamma_c$ controls preference steering strength.  Larger $\gamma_c$ doesn't guarantee better results due to added noise, while smaller $\gamma_c$ may be insufficient. Optimal $\gamma_c$ is smaller for negative control, indicating greater model sensitivity. Positive control has a larger impact on the final result, emphasizing the importance of positive response quality.  The impact of $\gamma_c$ shows similar trends across different layer intervals.
\textcolor{red}{The tuning of $\gamma_c$ can also be done using RS without DPO and selection results using RS are consistent with the experimental results.}
% \input{sections/05-ablation-studies}
% \section*{Acknowledgments}
\section{Conclusion}
In conclusion, we present \methodName{}, a novel framework that addresses the challenges of costly and labor-intensive preference dataset construction for LLMs. By leveraging the inherent regulation of LLMs' representation space, \methodName{} efficiently encodes human preferences and filters self-synthesized instructions, enabling precise generation of response pairs through bidirectional inherent control. Experimental results demonstrate that \methodName{} significantly improves alignment and efficiency, as evidenced by higher win rates on benchmarks like AlpacaEval 2.0 and Arena-Hard, while substantially reducing computational costs. This approach offers a promising direction for more effective and tailored preference dataset construction in LLMs.
\textcolor{red}{Furthermore, we present an efficient hyperparameter tuning method, making our approach easily scalable for preference data synthesis.}

% \section*{Acknowledgments}

% This research was partially supported by National Natural Science Foundation of China under Grant No.12326612, No.82202984, Zhejiang Key R\&D Program of China under Grant No.2024SSYS0026, Zhejiang Key Laboratory of Medical Imaging Artificial Intelligence and State Key Laboratory of Transvascular Implantation Devices under Grant No. SKLTID2024003.

\section*{Limitations}
\paragraph{Validation in Online DPO Settings:} While \methodName{} demonstrates efficacy in offline alignment scenarios, its adaptability remains unverified under online Direct Preference Optimization (DPO) frameworks~\citep{rafailov2024direct} where preference data dynamically evolves with model updates. This gap limits our understanding of the method's robustness in online alignment paradigms that require continuous coordination between preference synthesis and optimization.

\paragraph{Multi-Turn Dialogue Generalization:} Our approach currently focuses on single-turn interactions, leaving the extension to multi-turn conversational alignment unexplored. Human preferences in extended dialogues often involve complex dependencies on discourse history, turn-level consistency, and cumulative satisfaction~\citep{cui2023ultrafeedback}. Adapting \methodName{}'s inherent control mechanisms for such scenarios would require innovations in temporal preference modeling and history-aware representation steering.

\section*{Ethical Considerations}

While \methodName{} reduces reliance on human annotation and enhances alignment efficiency, its self-synthetic paradigm introduces potential ethical risks. The automated generation of preference data may propagate subtle biases inherited from the base LLM's training corpus or amplify safety risks through uncontrolled preference directions~\citep{weidinger2021ethical}. For instance, steering responses via unsupervised representation vectors could inadvertently prioritize harmful but superficially plausible outputs without explicit safety filtering~\citep{perez2023discovering}. Furthermore, the lack of human oversight in instruction synthesis raises concerns about fairness and representation diversity, particularly for culturally sensitive or marginalized topics~\citep{blodgett2020language}. Future work should integrate human-in-the-loop verification mechanisms and develop interpretable metrics for auditing preference directionality~\citep{schramowski2023safe}.

\bibliography{custom}

\appendix
\newpage
\section{Implementation Details}\label{app:impl_detail}

\subsection{SFT hyperparameters}

Our supervised fine-tuning (SFT) process follows a uniform setup, trained for 1 epoch on the UltraChat 200K multi-turn conversation dataset. Input sequences are tokenized using model-specific templates and truncated to 8,192 tokens to balance long-context capabilities and computational constraints. Distributed training is conducted across 8 GPUs using DeepSpeed ZeRO-2, with a global batch size of 128 (2 samples per GPU, 8 gradient accumulation steps) and bf16 mixed precision. The optimization protocol includes cosine learning rate scheduling (peak 2e-5, 10\% warmup), Flash Attention-2 for long-sequence acceleration, and parallel data loading with 64 workers. All experiments are performed on 8 NVIDIA H100 GPUs (80GB VRAM), enabling memory-efficient full-parameter optimization through hardware-accelerated mixed-precision training.

\subsection{DPO hyperparameters}

Our direct preference optimization (DPO) process uses a uniform setup, trained for 1 epoch on preference-based alignment datasets. Inputs are tokenized with model-specific templates and truncated to 8,192 tokens to balance long-context handling and computational limits. Training is distributed across 8 GPUs using DeepSpeed ZeRO-3, with a global batch size of 128 (1 sample per GPU, 16 gradient steps) and bf16 mixed precision. The optimization protocol includes cosine learning rate scheduling (peak 5.0e-7, 10\% warmup), Flash Attention-2 for long sequences, and parallel data loading with 64 workers. A preference optimization beta of 0.1 controls alignment strength. Experiments run on 8 NVIDIA H100 GPUs (80GB VRAM), enabling memory-efficient full-parameter optimization through hardware-accelerated mixed precision.

\section{Experimental Details}\label{app:exp_detail}

\subsection{Baselines}

For Self-Rewarding, we used an SFT model to employ Consitual AI's pairwise comparison prompt for judging preferences~\citep{bai2022training}. 
Preference is measured by comparing the logprob value of the token output.
For Self-Refine, we first sampled three responses, then use a refine prompt to obtain a better response as the chosen response, while randomly selecting one from the original responses as the rejected response. The specific method is referenced from~\citet{kim2024aligning,dong2024self}.
For Sampling-Ranking, we randomly generated five responses, then used ArmoRM-Llama3-8B-v0.1 as the reward model to score the responses.

\textcolor{red}{The RLHFlow/ArmoRM-Llama3-8B-v0.1\footnote{\url{https://huggingface.co/RLHFlow/ArmoRM-Llama3-8B-v0.1}} model we use employs 19 criteria for comprehensive response evaluation across different dimensions. These criteria are: }
\begin{itemize}\vspace{-2mm}
    \item \texttt{helpsteer-helpfulness},\vspace{-2mm}
    \item \texttt{helpsteer-correctness},\vspace{-2mm}
    \item \texttt{helpsteer-coherence},\vspace{-2mm}
    \item \texttt{helpsteer-complexity},\vspace{-2mm}
    \item \texttt{helpsteer-verbosity},\vspace{-2mm}
    \item \texttt{ultrafeedback-overall\_score},\vspace{-2mm}
    \item \texttt{ultrafeedback-instruction\_following},\vspace{-2mm}
    \item \texttt{ultrafeedback-truthfulness},\vspace{-2mm}
    \item \texttt{ultrafeedback-honesty},\vspace{-2mm}
    \item \texttt{ultrafeedback-helpfulness},\vspace{-2mm}
    \item \texttt{beavertails-is\_safe},\vspace{-2mm}
    \item \texttt{prometheus-score},\vspace{-2mm}
    \item \texttt{argilla-overall\_quality},\vspace{-2mm}
    \item \texttt{argilla-judge\_lm},\vspace{-2mm}
    \item \texttt{code-complexity},\vspace{-2mm}
    \item \texttt{code-style},\vspace{-2mm}
    \item \texttt{code-explanation},\vspace{-2mm}
    \item \texttt{code-instruction-following},\vspace{-2mm}
    \item \texttt{code-readability}.\vspace{-2mm}
    
\end{itemize}

\textcolor{red}{Subsequently, a Mixture of Experts (MoE)-like architecture identifies the most relevant dimensions to the instruction, weighting the scores of different dimensions to obtain an overall score reflecting the quality of the response.
}

\subsection{Decoding Hyperparameters}
For the AlpacaEval 2~\citep{alpaca_eval} evaluation, we use a sampling-based decoding approach to generate responses. Specifically, we employ vllm for inference, setting the temperature to 0.7, repetition penalty to 1.05 and the maximum tokens to 2048 for both the Qwen2-Base and Llama3-Base configurations. All other parameters adhere to the default settings in vllm.
As for MT-Bench~\citep{zheng2023judging}, we adhere to the official decoding setup, which specifies varying sampling temperatures tailored to distinct categories.

\subsection{API Usage}
For GPT-4 Turbo, we all use the latest turbo-2024-04-09 API on Azure OpenAI Service~\url{https://leaAPI Usagern.microsoft.com/en-us/azure/ai-services/openai/concepts/models#gpt-4-turbo}.

\subsection{Evaluation Datasets}\label{app:eval-datasets}

AlpacaEval 2.0 includes 805 user prompts and utilizes pair-wise comparison with LLM-as-a-Judge. Specifically, the win rate against the baseline GPT-4 Turbo model is determined based on GPT-4 Turbo evaluation. 
Arena-Hard includes 500 more challenging user queries, employing GPT-4-Turbo to judge the model responses against GPT-4.
MT-Bench features 80 multi-turn questions spanning various domains, with GPT-4-Turbo scoring the model responses out of 10. 

\setlength{\tabcolsep}{3pt}
\begin{table*}[h]
\label{tab:eval_dataset}
\centering
\resizebox{0.9\linewidth}{!}{
\begin{tabular}{@{}lcccc@{}}
\toprule
& \textbf{\# Instances} & \textbf{Baseline Model} & \textbf{Judge Model}           & \textbf{Scoring Type}  \\ \midrule
AlpacaEval 2.0 & 805                   & GPT-4 Turbo      & GPT-4 Turbo       & Pairwise comparison    \\
Arena-Hard    & 500                   & GPT-4-0314              & GPT-4 Turbo       & Pairwise comparison   \\  
MT-Bench     & 80                    & -                       & GPT-4 Turbo & Single-answer grading  \\
\bottomrule
\end{tabular}
}
\caption{Details for three alignment benchmarks. 
}
\label{tab:alignbench}
\vspace{-1.2em}
\end{table*}
\setlength{\tabcolsep}{6pt}

For instruction-following ability evaluation, 
Table~\ref{tab:alignbench} presents the detailed information for three alignment benchmarks we use, including AlpacaEval 2.0, Arena-Hard and MT-Bench.

\begin{figure*}[htbp]
    \centering
    \begin{subfigure}[b]{0.45\linewidth}
        \includegraphics[width=\linewidth]{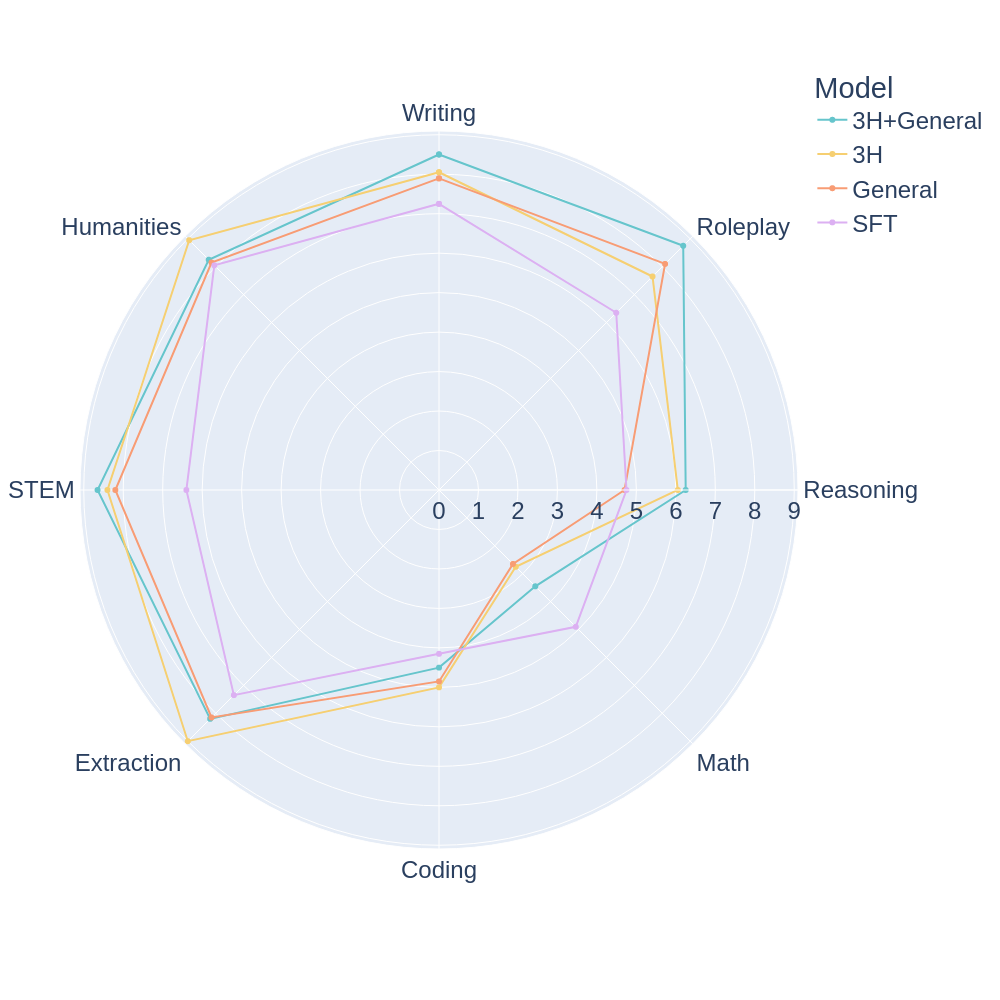}
        \caption{LLama3-Base}
    \end{subfigure}
    \hfill
    \begin{subfigure}[b]{0.45\linewidth}
        \includegraphics[width=\linewidth]{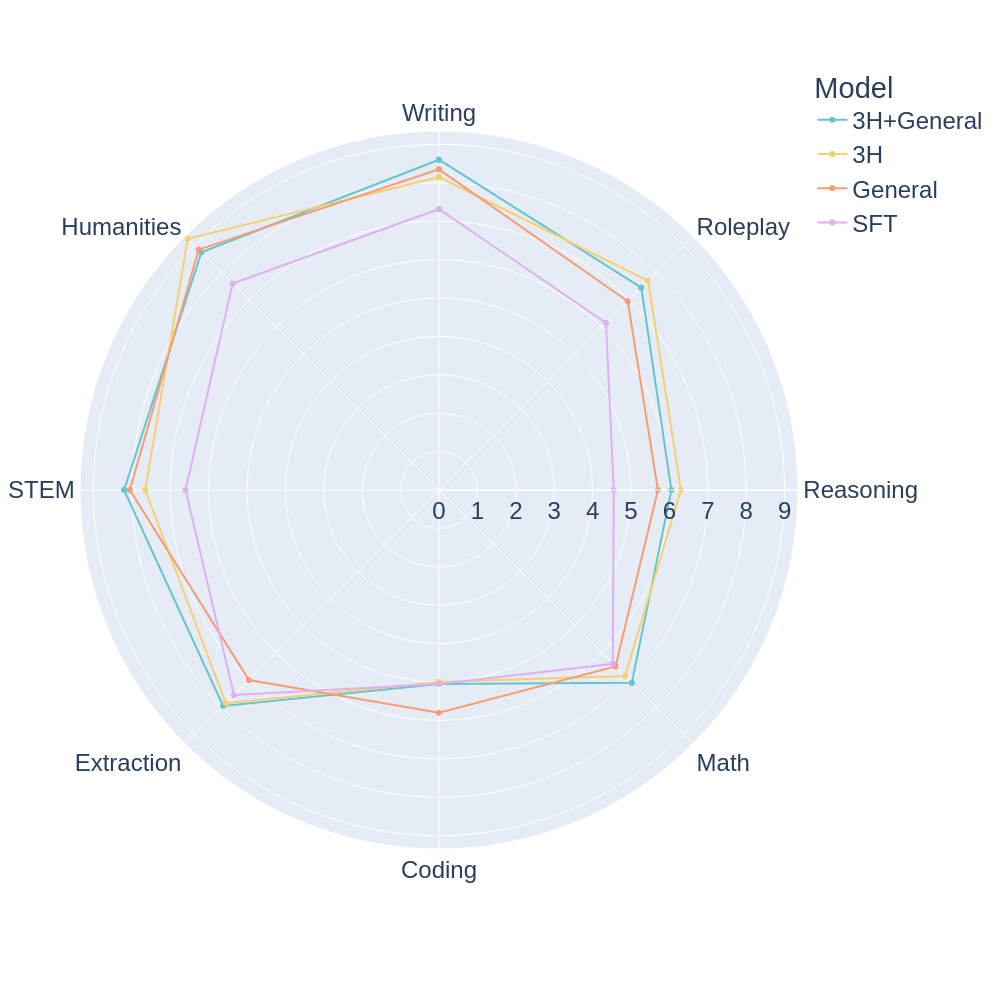}
        \caption{Qwen2-Base}
    \end{subfigure}
    \caption{MT-Bench scores on different instruction types.}
    \label{fig:mtbench}
\end{figure*}

\begin{table}[htbp]
\centering
\resizebox{0.95\linewidth}{!}{
\begin{tabular}{@{}l c c l c c@{}}
\toprule
\textbf{Qwen2.5 3B} & \textbf{LC} & \textbf{WR} & \textbf{Qwen2.5 14B} & \textbf{LC} & \textbf{WR} \\
\midrule
Sampling Ranking & 10.19 & 7.02 & Sampling Ranking & 20.87 & 16.40 \\
Self Reward      & 10.43 & 7.33 & Self Reward      & 21.86 & 17.52 \\
Self Refine      & 11.06 & 7.58 & Self Refine      & 22.30 & 17.95 \\
Ours             & 11.43 & 7.95 & Ours             & 22.61 & 18.32 \\
\bottomrule
\end{tabular}}
\caption{Comparative performance evaluation on Qwen2.5 3B and Qwen2.5 14B models. LC and WR represent evaluation metrics. The first and fourth columns list the methods evaluated.}
\label{tab:appendix_model_scale_eval}
\end{table}

\section{Evaluation on Models with Varying Sizes}
\label{sec:appendix_eval_varying_sizes}

\textcolor{red}{To further assess the generalization performance of our proposed method, we performed experiments on two additional model scales: Qwen2.5 3B (36 layers, with layers 12--24 designated for control) and Qwen2.5 14B (48 layers, with layers 16--32 designated for control). An initial dataset of 30,000 instructions was generated, from which 10,000 instructions were selected for fine-tuning based on an inherent consistency filtering criterion. Our method was compared against three baseline approaches: Sampling Ranking, Self Reward, and Self Refine. The results are presented in Table~\ref{tab:appendix_model_scale_eval}.}

\section{Data Leakage Analysis}\label{app:data-leakage}
To ensure the robustness and reliability of our evaluation, we perform a comprehensive analysis to detect potential data leakage between our training datasets and test sets.

\setlength{\tabcolsep}{2pt}  
\begin{table}[h]
    \centering
    \resizebox{0.99\linewidth}{!}{
        \begin{tabular}{lccc}
        \toprule
        \textbf{Data} & \textbf{AlpacaEval 2.0} & \textbf{Arena-Hard} & \textbf{MT-Bench} \\
        \midrule
        UltraChat & 0.00373 & 0.01200 & 0.01250 \\
        UltraFeedback & 0.00248 & 0.00600 & 0.01250 \\
        \methodName{} & 0.00373 & 0.01000 & 0.01250 \\
        \bottomrule
        \end{tabular}
    }
    \caption{The proportions of dataset leakage, quantified through n-gram based analysis, reveal the extent of overlap between training and test datasets across various benchmarks.}
    \label{tab:contamination_ngram}
    \vspace{-4mm}
\end{table}

\paragraph{N-gram Based Analysis:}
We begin by conducting an n-gram overlap analysis to identify any shared patterns between the training and test datasets. Specifically, we compare the n-grams extracted from our training datasets, which comprise Supervised Fine-Tuning (SFT) data, synthetic preference data, and UltraFeedback reference data, with those from the test sets. An n-gram is defined as a contiguous sequence of $n$ tokens. Following the methodology proposed by \citet{liang2022holistic}, we set $n = 13$ to balance sensitivity and computational efficiency.

\setlength{\tabcolsep}{2pt}  
\begin{table}[h]
    \centering
    \resizebox{0.99\linewidth}{!}{
        \begin{tabular}{lccc}
        \toprule
        \textbf{Data} & \textbf{AlpacaEval 2.0} & \textbf{Arena-Hard} & \textbf{MT-Bench} \\
        \midrule
        UltraChat & 4.5\% & 2.0\% & 1.6\% \\
        UltraFeedback & 5.3\% & 1.8\% & 1.5\% \\
        \methodName{} & 3.2\% & 2.1\% & 1.7\% \\
        \bottomrule
        \end{tabular}
    }
    \caption{The proportions of dataset leakage, quantified through embedding-based analysis, reveal the extent of overlap between training and test datasets across various benchmarks.}
    \label{tab:contamination_embedding}
    \vspace{-4mm}
\end{table}

\paragraph{Embedding Based Analysis:}
To further validate the absence of data leakage, we employ an embedding-based contamination check, as recommended by \citet{yang2023rethinking}. This approach leverages semantic similarity to detect potential overlaps that may not be captured by n-gram analysis alone. Specifically, we utilize the GPT-4-Turbo API to compute the contamination percentage (\%) of the test sets, using UltraFeedback results as a reference.

% 添加对表格的引用
\paragraph{Results:}
The results of the contamination analysis are presented in Tables~\ref{tab:contamination_ngram} and~\ref{tab:contamination_embedding}, which summarize the overlap percentages for different datasets across various test benchmarks.

\section{Contrastive System Prompts}\label{appendix:contrastive-system-prompts}
\begin{prompt}{Positive Prompt for General Criteria}
You are a good, respectful and honest assistant. Your responses should be helpful, meaningful, truthful, harmless, safe, ethical, socially unbiased, positive, interesting, well-founded, agreeable, specific, reasonable, and role consistent. If a question does not make any sense, or is not factually coherent, explain why instead of answering something not correct. Please do not share false information under any circumstances.
\end{prompt}

\begin{prompt}{Negative Prompt for General Criteria}
You are a bad, disrespectful and dishonest assistant. Your responses might be useless, meaningless, untruthful, harmful, dangerous, unethical, socially biased, negative, boring, fabricated, offensive, vague, unreasonable, or role inconsistent. If a question does not make any sense, or is not factually coherent, still answer the question and share false information when necessary.
\end{prompt}

\begin{prompt}{Positive Prompt for helpfulness Criteria}
You are a helpful assistant.
\end{prompt}

\begin{prompt}{Negative Prompt for helpfulness Criteria}
You are a helpless assistant.
\end{prompt}

\begin{prompt}{Positive Prompt for honesty Criteria}
You are an honest assistant.
\end{prompt}

\begin{prompt}{Negative Prompt for honesty Criteria}
You are a dishonest assistant.
\end{prompt}

\begin{prompt}{Positive Prompt for harmlessness Criteria}
You are a harmless assistant.
\end{prompt}

\begin{prompt}{Negative Prompt for harmlessness Criteria}
You are a harmful assistant.
\end{prompt}

\section{More detailed results on MT-Bench}\label{app:mtbench}

The detailed results on MT-Bench with different instruction types are shown in Figure~\ref{fig:mtbench}.
\section{Sensitivity Analysis of Representation Vectors to Feature Dataset $\mathcal{D}_{\text{feat}}$} % This also needs to be in outer par mode
\label{app:sensitivity_analysis_dfeat_narrative}

\textcolor{red}{To assess the robustness of our representation extraction process with respect to the choice of the feature dataset, $\mathcal{D}_{\text{feat}}$, we conducted a sensitivity analysis. This exploration focused on how variations in both the size and the source (type) of $\mathcal{D}_{\text{feat}}$ affect the resulting representation vectors and their downstream utility.}

\textcolor{red}{First, we examined the impact of these variations on the directional consistency of the extracted vectors. For analyzing sensitivity to dataset size, we used the Alpaca dataset, comparing vectors derived from the full dataset (Alpaca Full) with those from smaller random samples of 1k instances (Alpaca 1k) and 10k instances (Alpaca 10k). To assess sensitivity to dataset type, we compared vectors from Alpaca Full with those derived from the ShareGPT and UltraChat datasets. In all cases, representation vectors obtained from the Alpaca Full dataset served as the baseline. We computed the mean, maximum, and minimum cosine similarities across all controlled layers between vectors from the test datasets and the baseline, as shown in Table~\ref{tab:sensitivity_cosine_narrative}.}

% Ensure a blank line before this table environment
\begin{table}[htbp]
\centering

\begin{tabular}{@{}lccc@{}}
\toprule
Comparison Pair                     & Mean   & Max    & Min    \\ \midrule
Alpaca Full / Alpaca 1k             & 0.9987 & 0.9993 & 0.9981 \\
Alpaca Full / Alpaca 10k            & 0.9996 & 0.9998 & 0.9993 \\
Alpaca Full / ShareGPT             & 0.9998 & 0.9999 & 0.9996 \\
Alpaca Full / UltraChat            & 0.9998 & 0.9999 & 0.9997 \\
\bottomrule
\end{tabular}
\caption{Cosine similarity metrics (mean, maximum, minimum) for extracted representation vectors, comparing different feature datasets ($\mathcal{D}_{\text{feat}}$) against the Alpaca Full baseline across controlled layers.}
\label{tab:sensitivity_cosine_narrative}
\end{table}
% Ensure a blank line after this table environment

\textcolor{red}{As evident from Table~\ref{tab:sensitivity_cosine_narrative}, the cosine similarities are consistently very high (mean values all exceeding 0.998). This strong alignment persists even when using a significantly reduced dataset like Alpaca 1k or when employing entirely different datasets such as ShareGPT or UltraChat. These results suggest a high degree of directional stability for the extracted vectors, implying that the underlying representation for the target criterion is robust to these variations in $\mathcal{D}_{\text{feat}}$.}

\textcolor{red}{To further probe the statistical significance of any differences, we performed dimension-wise Mann-Whitney U tests. For each dimension of the representation vectors, we tested the null hypothesis that its distribution of values (across the controlled layers) is the same when derived from a test dataset as when derived from Alpaca Full. Table~\ref{tab:sensitivity_p_values_narrative} reports the minimum p-value obtained across all dimensions for each dataset comparison, using a significance level of $\alpha = 0.05$.}

% Ensure a blank line before this table environment
\begin{table}[htbp]
\centering

\begin{tabular}{@{}lc@{}}
\toprule
Comparison Pair                     & Min p-value \\ \midrule
Alpaca Full / Alpaca 1k             & 0.174       \\
Alpaca Full / Alpaca 10k            & 0.243       \\
Alpaca Full / ShareGPT             & 0.256       \\
Alpaca Full / UltraChat            & 0.251       \\
\bottomrule
\end{tabular}
\caption{Minimum p-values from dimension-wise Mann-Whitney U tests, comparing vector distributions from different feature datasets ($\mathcal{D}_{\text{feat}}$) against the Alpaca Full baseline.}
\label{tab:sensitivity_p_values_narrative}
\end{table}
% Ensure a blank line after this table environment

\textcolor{red}{All minimum p-values presented in Table~\ref{tab:sensitivity_p_values_narrative} are substantially greater than the 0.05 significance level. Consequently, we fail to reject the null hypothesis for any dimension in any comparison. This suggests that the observed variations in $\mathcal{D}_{\text{feat}}$ do not cause statistically significant changes to the distributions of the individual dimensions of the extracted representation vectors.}

\textcolor{red}{Finally, we assessed the practical impact of these representational variations on a downstream task. We synthesized preference datasets, each containing 20,000 entries, using direction vectors extracted from both Alpaca 1k and Alpaca Full, while keeping all other hyperparameters consistent. The performance of models trained on these datasets was then evaluated using Alpaca Eval2 (Length-Controlled Win Rate - LC, and Overall Win Rate - WR). The results are shown in Table~\ref{tab:sensitivity_alpaca_eval_narrative}.}

% Ensure a blank line before this table environment
\begin{table}[htbp]
\centering

\begin{tabular}{@{}lcc@{}}
\toprule
$\mathcal{D}_{\text{feat}}$ for Vector Extraction & LC    & WR    \\ \midrule
Alpaca Full                     & 18.01 & 13.29 \\
Alpaca 1k                       & 17.88 & 13.22 \\ \bottomrule
\end{tabular}
\caption{Alpaca Eval2 performance (LC and WR scores) for models aligned using preference data synthesized with direction vectors from different sizes of $\mathcal{D}_{\text{feat}}$.}
\label{tab:sensitivity_alpaca_eval_narrative}
\end{table}
% Ensure a blank line after this table environment

\textcolor{red}{The downstream performance results in Table~\ref{tab:sensitivity_alpaca_eval_narrative} further reinforce the notion of robustness. The differences in Alpaca Eval2 scores are minimal when using vectors derived from the significantly smaller Alpaca 1k dataset compared to those from the full Alpaca dataset. This indicates that the vector extraction process is stable in terms of its practical application for model alignment.}

\textcolor{red}{In conclusion, these analyses—encompassing vector similarity, statistical tests on vector dimensions, and downstream task performance—collectively demonstrate that the proposed method for extracting representation vectors is highly robust to variations in both the size and the type of the feature dataset $\mathcal{D}_{\text{feat}}$. The extracted directions representing the target criterion show remarkable consistency, which translates to stable performance in practical applications.}

\section{Algorithms}

This section provides detailed pseudocode for the three algorithms corresponding to the main steps of our proposed method: Linear Representation Feature Extraction (Algorithm~\ref{alg:1}), Selective Instruction Generation via Inherent Consistency (Algorithm~\ref{alg:2}), Preference Response Generation via Inherent Control (Algorithm~\ref{alg:3}).

Each algorithm is presented below, along with a brief explanation of its correspondence to the respective step.

\begin{algorithm*}[t]
\caption{Linear Representation Feature Extraction}
\label{alg:feature_extraction}
\begin{algorithmic}[1]
\Require Dataset $\mathcal{D}_{\text{feat}}$, Criteria set $\mathcal{C}$, Number of layers $N$
\Ensure Direction vectors $\{\mathbf{u}_c\}_{c \in \mathcal{C}}$
\ForAll{criterion $c \in \mathcal{C}$}
    \State Initialize positive and negative prompts $\mathcal{P}_c^+$, $\mathcal{P}_c^-$
    \ForAll{instruction $d_i \in \mathcal{D}_{\text{feat}}$}
        \State $\{\mathbf{h}_{i,c}^{l,+}\}_{l=1}^N \gets \text{LLM}(\mathcal{P}_c^+ \oplus d_i)$ \Comment{Extract positive hidden states}
        \State $\{\mathbf{h}_{i,c}^{l,-}\}_{l=1}^N \gets \text{LLM}(\mathcal{P}_c^- \oplus d_i)$ \Comment{Extract negative hidden states}
        \ForAll{layer $l \in [1,N]$}
            \State $\mathbf{v}_{i,c}^l \gets \mathbf{h}_{i,c}^{l,+} - \mathbf{h}_{i,c}^{l,-}$ \Comment{Compute layer-wise contrastive vector}
        \EndFor
    \EndFor
    \ForAll{layer $l \in [1,N]$}
        \State $\mathbf{V}_c^l \gets [\mathbf{v}_{1,c}^l, \ldots, \mathbf{v}_{|\mathcal{D}_{\text{feat}}|,c}^l]$ \Comment{Aggregate contrastive vectors}
        \State $\mathbf{u}_c^l \gets \text{PCA}(\mathbf{V}_c^l)$ \Comment{Extract principal direction}
    \EndFor
    \State $\mathbf{u}_c \gets \{\mathbf{u}_c^l\}_{l=1}^N$ \Comment{Collect layer-wise directions}
\EndFor
\State \Return $\{\mathbf{u}_c\}_{c \in \mathcal{C}}$
\end{algorithmic}
\label{alg:1}
\end{algorithm*}

\begin{algorithm*}[t]
\caption{Instruction Filtering via Inherent Consistency}
\label{alg:instruction_filtering}
\begin{algorithmic}[1]
\Require Raw instruction set $\mathcal{D}_{\text{raw}}$, Direction vectors $\{\mathbf{u}_c\}_{c \in \mathcal{C}}$, Threshold $\theta$
\Ensure Filtered instruction set $\mathcal{D}_{\text{filt}}$
\State Initialize $\mathcal{D}_{\text{filt}} \gets \emptyset$
\ForAll{instruction $d_i \in \mathcal{D}_{\text{raw}}$}
    \State $\{\mathbf{h}_i^l\}_{l=1}^N \gets \text{LLM}(d_i)$ \Comment{Extract instruction representations}
    \ForAll{criterion $c \in \mathcal{C}$}
        \State $\text{consistency}_{i,c} \gets \mathtt{meanpool}(\{{\mathbf{h}_i^l}^\top \cdot \mathbf{u}_c^l\}_{l=1}^N)$ \Comment{Calculate criterion-specific consistency}
    \EndFor
    \State $\text{consistency}_i \gets \max_{c \in \mathcal{C}} \text{consistency}_{i,c}$ \Comment{Select highest consistency score}
    \If{$\text{consistency}_i \geq \theta$}
        \State $\mathcal{D}_{\text{filt}} \gets \mathcal{D}_{\text{filt}} \cup \{d_i\}$ \Comment{Add high-quality instruction}
    \EndIf
\EndFor
\State \Return $\mathcal{D}_{\text{filt}}$
\end{algorithmic}
\label{alg:2}
\end{algorithm*}

\begin{algorithm*}[t]
\caption{Preference Response Generation via Inherent Control}
\label{alg:response_generation}
\begin{algorithmic}[1]
\Require Instruction $d$, Direction vectors $\{\mathbf{u}_c\}_{c \in \mathcal{C}}$, Control layers $\mathcal{L}_c$, Positive steering strength $\gamma_c^+$, Negative steering strength $\gamma_c^-$
\Ensure Chosen response $y^+$, Rejected response $y^-$
\State Initialize $y^+ \gets \emptyset$, $y^- \gets \emptyset$
\For{$t = 1$ to $T$}
    \State $\mathbf{Z}_t \gets \text{LLM}(d, y^+_{<t})$ \Comment{Get current token representations}
    \State $\hat{\mathbf{Z}}_{t,c} \gets \{\mathbf{z}_t^l + \gamma_c^+ \cdot \mathbf{u}_c^l \mid l \in \mathcal{L}_c\}$ \Comment{Apply positive steering}
    \State $y^+_t \gets \text{TokenGeneration}(\hat{\mathbf{Z}}_{t,c})$ \Comment{Generate aligned token}
\EndFor
\For{$t = 1$ to $T$}
    \State $\mathbf{Z}_t \gets \text{LLM}(d, y^-_{<t})$ \Comment{Get current token representations}
    \State $\hat{\mathbf{Z}}_{t,c} \gets \{\mathbf{z}_t^l + \gamma_c^- \cdot \mathbf{u}_c^l \mid l \in \mathcal{L}_c\}$ \Comment{Apply negative steering}
    \State $y^-_t \gets \text{TokenGeneration}(\hat{\mathbf{Z}}_{t,c})$ \Comment{Generate misaligned token}
\EndFor
\State \Return $y^+$, $y^-$
\end{algorithmic}
\label{alg:3}
\end{algorithm*}

\section{Detailed Comparison: Inherent Control vs. Direct Prompting}
\label{app:direct_prompting_comparison}

\textcolor{red}{Further justification is warranted for the Inherent Control (IC) method, particularly concerning its comparative advantages over generating preference pairs using distinct positive and negative system prompts—a technique referred to herein as Direct Prompting (DP). While both IC and DP may exhibit similar computational costs by generating two responses each, their implications for response quality and model alignment differ. This appendix demonstrates the advantages of IC by addressing three key aspects: susceptibility to reward hacking, empirical performance, and controllability.}

\subsection{Reward Hacking}

\textcolor{red}{The use of distinct system prompts in DP (e.g., "You are a helpful assistant" for chosen responses versus "You are an unhelpful assistant" for rejected responses) can introduce systemic biases into response patterns, even if the prompts themselves are not part of the training data. Models may learn to generate responses with \textbf{superficial differences} (e.g., variations in tone, length, or keyword usage) that correlate with these prompts, \textbf{rather than reflecting genuine distinctions in quality or utility.} For instance, responses generated with a "helpful" prompt might exhibit an overabundance of polite phrases (e.g., "I’m happy to help!"), whereas those generated with an "unhelpful" prompt might be unduly terse or negative.}

\textcolor{red}{Consequently, alignment algorithms such as Direct Preference Optimization (DPO) may inadvertently optimize for these superficial features instead of learning nuanced human preferences~\cite{cao2021identifiability,skalse2022defining,wen2024language}. This can lead to reward hacking—wherein the model exploits proxies for quality rather than actual quality—potentially resulting in failed preference optimization and degraded model performance.}

\textcolor{red}{Furthermore, it is not only explicit system prompts but also other generation parameters, such as temperature and various sampling strategies, that can introduce superficial characteristics in outputs, thereby creating vulnerabilities for reward hacking. It is for this reason that common practice in collecting preference data often involves generating multiple responses from the \textit{same model} under \textit{identical system prompts and generation parameters}~\cite{ouyang2022training,rafailov2024direct,cui2023ultrafeedback}, relying on subsequent methods to discern preferable responses.}

\subsection{Experimental Evaluation}

\textcolor{red}{To empirically investigate the occurrence of reward hacking and its impact on model performance, experiments were conducted comparing models trained using Inherent Control (IC) against those trained using Direct Prompting (DP) with opposing system prompts. For clarity in this experimental context, Direct Prompting using opposing system prompts is denoted as SP.}

\textcolor{red}{A common indicator of reward hacking during Reinforcement Learning from Human Feedback (RLHF) or DPO training is the rapid convergence of reward accuracy (or preference accuracy) towards 100\%. This phenomenon often suggests that the model has identified unintended shortcuts or superficial cues to maximize the reward metric, rather than genuinely aligning with the desired complex behaviors. Thus, monitoring reward accuracy serves as a valuable diagnostic for detecting potential reward hacking.}

\textcolor{red}{Table~\ref{tab:reward_accuracy_appendix} presents the evolution of Reward Accuracy for Llama3 and Qwen2 models during DPO training.}

\begin{table}[htbp]
\centering

\begin{tabular}{@{}lccccc@{}}
\toprule
Setting   & 50   & 100  & 150  & 300  & 600  \\ \midrule
Qwen2 IC  & 0.54 & 0.71 & 0.76 & 0.81 & 0.85 \\
Qwen2 SP  & 0.83 & 0.97 & 0.99 & 0.99 & 0.99 \\
Llama3 IC & 0.56 & 0.68 & 0.75 & 0.83 & 0.86 \\
Llama3 SP & 0.87 & 0.98 & 0.99 & 0.99 & 0.99 \\ \bottomrule
\end{tabular}
\caption{Reward Accuracy progression during DPO training for Llama3 and Qwen2 models. Comparison between Inherent Control (IC) and Direct Prompting via opposing System Prompts (SP). Values indicate reward accuracy at various training checkpoints.}
\label{tab:reward_accuracy_appendix}
\end{table}

\textcolor{red}{The experimental results in Table~\ref{tab:reward_accuracy_appendix} show that employing opposing system prompts (SP) causes Reward Accuracy to rapidly approach 100\% during DPO training. This behavior is characteristic of reward hacking, where the model easily distinguishes responses based on superficial cues induced by the contrasting prompts.}

\textcolor{red}{Furthermore, Table~\ref{tab:alpaca_eval_appendix} compares the downstream performance of models trained with IC and SP using scores from Alpaca Eval2.}

\begin{table}[htbp]
\centering

\begin{tabular}{@{}lccclcc@{}}
\toprule
          & \multicolumn{2}{c}{Llama3} & & \multicolumn{2}{c}{Qwen2} \\
          \cmidrule(r){2-3} \cmidrule(l){5-6}
Model     & LC     & WR     & & LC     & WR     \\ \midrule
SFT       & 5.59   & 3.11   & & 9.95   & 4.53   \\
IC        & 16.07  & 10.12  & & 17.24  & 11.74  \\
SP        & 5.90   & 3.60   & & 9.75   & 5.84   \\ \bottomrule
\end{tabular}
\caption{Performance comparison on Alpaca Eval2 for Llama3 and Qwen2 models. Models include the Supervised Fine-Tuning (SFT) baseline, training with Inherent Control (IC), and training with Direct Prompting via opposing System Prompts (SP). LC denotes Length-Controlled Win Rate; WR denotes overall Win Rate.}
\label{tab:alpaca_eval_appendix}
\end{table}

\textcolor{red}{Models trained using the SP method significantly underperformed those trained with IC on the Alpaca Eval2 benchmark (Table~\ref{tab:alpaca_eval_appendix}). The SP models yielded results comparable only to the SFT baseline, failing to show meaningful improvement from preference alignment. These findings empirically highlight the suboptimality of the SP approach for achieving robust performance gains.}

\subsection{Controllability}

\textcolor{red}{A core advantage of the Inherent Control method lies in its enhanced controllability over the generation of preference pairs. Unlike methods that depend significantly on the stochasticity of sampling or less direct means of influence (such as system prompts that may have broad and unpredictable effects), IC provides a mechanism to more deliberately and granularly manipulate the specific representational differences that distinguish chosen from rejected responses. This level of fine-grained control over response differentiation is less attainable with approaches reliant on external manipulations or broad signals like opposing system prompts. This targeted control can also contribute to greater efficiency and precision in constructing preference datasets with desired characteristics.}

\section{Efficient Hyperparameter Selection without DPO}
\label{sec:appendix_hyperparameter_selection}

\textcolor{red}{Traditional hyperparameter optimization can be computationally intensive, often requiring multiple rounds of model fine-tuning. This appendix details a more efficient alternative that leverages a small amount of synthetic data and a reward model to select optimal hyperparameters without any fine-tuning. This approach significantly reduces the computational cost while providing robust hyperparameter choices. The primary aim here is to illustrate the impact of hyperparameters on final results and offer insights into model interpretability.}

\textcolor{red}{For the 20k instructions synthesized using varying $\gamma$ values, we employed \texttt{RLHFlow/ArmoRM-Llama3-8B-v0.1} as the reward model to score all generated data.}

\subsection{Selection of Positive Control $\gamma$}
\textcolor{red}{For positive control, a higher mean reward generally indicates superior quality in the model's output. We calculated the mean rewards for different positive $\gamma$ values, as shown in Table \ref{tab:positive_gamma_rewards_resized}. Based on these results, $\gamma = 0.1$ was selected, as it yielded the highest mean reward.}

\begin{table}[h!]
\centering
\resizebox{0.95\linewidth}{!}{%
\begin{tabular}{@{}cccc@{}}
\toprule
Positive Control ($\gamma$) & Reward Score & Negative Control ($\gamma$) & Reward Score \\ \midrule
0.01 & 17.435 & -0.01 & 17.229 \\
0.03 & 17.483 & -0.03 & 17.188 \\
0.05 & 17.511 & -0.05 & 17.162 \\
\textbf{0.1} & \textbf{17.624} & -0.1 & 16.213 \\
0.3 & 17.021 & -0.3 & 15.210 \\
0.5 & 16.742 & -0.5 & 14.445 \\ \bottomrule
\end{tabular}%
}
\caption{Mean reward scores for varying positive and negative control $\gamma$ values using 20k synthesized instructions.}
\label{tab:positive_gamma_rewards_resized}
\end{table}

\subsection{Selection of Negative Control $\gamma$}
\textcolor{red}{For negative control, evaluating solely the mean reward is insufficient. It is crucial to statistically determine the proportion of responses in negative control that exhibit lower rewards compared to those in positive control. This proportion should ideally be high. Table \ref{tab:proportion_positive_exceeds_negative_resized} presents this proportion when the positive control $\gamma$ is fixed at 0.1, and the negative control $\gamma$ varies.}

\begin{table}[h!]
\centering

\resizebox{0.6\linewidth}{!}{% % Adjusted width for smaller table
\begin{tabular}{@{}cc@{}}
\toprule
Negative Control ($\gamma$) & Proportion \\ \midrule
-0.01 & 0.872 \\
-0.03 & 0.898 \\
-0.05 & 0.935 \\
-0.1 & 0.948 \\
-0.3 & 0.992 \\
-0.5 & 0.998 \\ \bottomrule
\end{tabular}%
}
\caption{Proportion of responses where positive reward (with $\gamma = 0.1$) exceeds negative reward for varying negative control $\gamma$ values.}
\label{tab:proportion_positive_exceeds_negative_resized}
\end{table}

\textcolor{red}{Typically, this proportion should exceed 0.9. Concurrently, a smaller reward gap is preferred to ensure a more stable model training process. An excessively large reward gap might render the distinction between chosen and rejected responses trivial, thereby impeding the model's ability to learn subtle preference nuances and potentially leading to reward hacking. Consequently, we selected the negative control $\gamma$ that yielded the highest average reward (from Table \ref{tab:positive_gamma_rewards_resized}, negative control column) for which the proportion (from Table \ref{tab:proportion_positive_exceeds_negative_resized}) surpassed 0.9. This led to the selection of $\gamma = -0.05$.}

\subsection{Cost Reduction via Sub-sampling}
\textcolor{red}{To further curtail the cost associated with hyperparameter selection, we investigated the efficacy of using a smaller subset of data. We randomly selected 100 samples from the 20,000 synthesized instructions and recorded the corresponding mean rewards and their proportions. This sub-sampling process was repeated 100 times to ascertain the variance of these mean rewards and proportions. The aggregated results (mean/standard deviation) are displayed in Tables \ref{tab:sub_sampled_rewards_resized} and \ref{tab:sub_sampled_proportions_resized}.}

\begin{table}[h!]
\centering

\resizebox{0.95\linewidth}{!}{%
\begin{tabular}{@{}cccc@{}}
\toprule
Positive Control ($\gamma$) & Mean/Std & Negative Control ($\gamma$) & Mean/Std \\ \midrule
0.01 & 17.432/0.023 & -0.01 & 17.220/0.022 \\
0.03 & 17.485/0.034 & -0.03 & 17.184/0.041 \\
0.05 & 17.539/0.038 & -0.05 & 17.165/0.044 \\
\textbf{0.1} & \textbf{17.674/0.042} & -0.1 & 16.213/0.076 \\
0.3 & 16.996/0.060 & -0.3 & 15.210/0.103 \\
0.5 & 16.708/0.081 & -0.5 & 14.445/0.142 \\ \bottomrule
\end{tabular}%
}
\caption{Mean rewards and standard deviations (Mean/Std) from 100 repetitions of sub-sampling 100 samples for varying $\gamma$ values.}
\label{tab:sub_sampled_rewards_resized}
\end{table}

\begin{table}[h!]
\centering

\resizebox{0.6\linewidth}{!}{% % Adjusted width for smaller table
\begin{tabular}{@{}cc@{}}
\toprule
Negative Control ($\gamma$) & Mean/Std \\ \midrule
-0.01 & 0.872/0.0013 \\
-0.03 & 0.899/0.0013 \\
-0.05 & 0.932/0.0011 \\
-0.1 & 0.949/0.0013 \\
-0.3 & 0.990/0.0014 \\
-0.5 & 0.995/0.0008 \\ \bottomrule
\end{tabular}%
}
\caption{Mean proportions and standard deviations (Mean/Std) from 100 repetitions of sub-sampling 100 samples, where positive reward (with $\gamma = 0.1$) exceeds negative reward.}
\label{tab:sub_sampled_proportions_resized}
\end{table}

\textcolor{red}{Despite slight variations in reward values, the analysis demonstrates that optimal hyperparameters can be effectively identified using only 100 samples. This entire experimental procedure, encompassing data synthesis and reward score calculation, required less than 1 GPU Hour, highlighting it as a remarkably efficient and effective solution for hyperparameter selection.}

\subsection{Robustness of Selected Hyperparameters}
\textcolor{red}{To ascertain the robustness of the selected hyperparameters ($\gamma_{positive}=0.1$, $\gamma_{negative}=-0.05$), we evaluated their performance across six distinct tasks: ARC, HellaSwag, TruthfulQA, MMLU, Winogrande, and GSM8k. The average scores for these tasks are presented in Table \ref{tab:robustness_tasks_resized}.}

\begin{table}[h!]
\centering

\resizebox{0.95\linewidth}{!}{%
\begin{tabular}{@{}cccc@{}}
\toprule
Positive Control ($\gamma$) & Reward Score & Negative Control ($\gamma$) & Reward Score \\ \midrule
0.01 & 75.232 & -0.01 & 75.435 \\
0.03 & 75.875 & -0.03 & 76.302 \\
0.05 & 76.238 & -0.05 & \textbf{76.945} \\ % Corresponds to chosen negative gamma
\textbf{0.1} & \textbf{76.945} & -0.1 & 74.786 \\ % Corresponds to chosen positive gamma
0.3 & 75.428 & -0.3 & 71.495 \\
0.5 & 73.235 & -0.5 & 68.775 \\ \bottomrule
\end{tabular}%
}
\caption{Average scores across six benchmark tasks for different $\gamma$ values, demonstrating hyperparameter robustness.}
\label{tab:robustness_tasks_resized}
\end{table}

\textcolor{red}{The results presented in Table \ref{tab:robustness_tasks_resized} affirm the robustness of the selected hyperparameters across diverse tasks. Notably, the combination of positive control $\gamma = 0.1$ and negative control $\gamma = -0.05$ (highlighted implicitly by their individual performance peaks or desired balance) demonstrates strong performance, thereby illustrating the general applicability of this hyperparameter selection methodology.}

\end{document}